\documentclass[10pt, a4paper]{article}

\usepackage[final]{lrec2026}
\usepackage{amsmath}
\usepackage{booktabs}
\usepackage{graphicx}

\usepackage{bidi}
\newfontfamily\arabicfont[Script=Arabic,Path=./]{NotoNaskhArabic-Regular.ttf}
\newcommand{\ar}[1]{{\arabicfont\RL{#1}}}

\title{Thaka at KSAA-2026 Task 2: Regularized Fine-Tuning\\for Arabic Speech Diacritization}

\name{Meshal Alamr, Hassan Alqaeri, Abdullah Aldahlawi}

\address{Thaka, Advanced AI and Information Technology \\
         Riyadh, Saudi Arabia \\
         \{m.alamr, h.alqaeril, aldahlawi\}@thakaait.net}

\abstract{
We describe the winning system for Task~2 of the KSAA-2026 Shared Task on Arabic Speech Dictation with Automatic Diacritization \citep{ksaa2026task}.
The task requires producing fully diacritized Arabic text from speech audio and undiacritized transcripts, with only 2{,}327 training samples available and no external data permitted.
Our system fine-tunes CATT-Whisper \citep{ghannam2025catt-whisper}, a character-level multimodal model combining a pretrained CATT text encoder with a frozen Whisper speech encoder.
The key to our approach is training regularization: R-Drop consistency regularization, Optuna-optimized hyperparameters with high weight decay, and Focal Loss.
At inference, we average 200 stochastic forward passes across four model checkpoints using Monte Carlo Dropout at the softmax probability level.
The system achieves \textbf{23.26\% WER} on the primary leaderboard metric (with case endings, including no-diacritic positions), placing 1st among all participants.
\\ \newline \Keywords{Arabic diacritization, multimodal, speech processing, regularization, shared task, KSAA-2026}}

\begin{document}

\maketitleabstract

\section{Introduction}
\label{sec:intro}

Arabic text is typically written without diacritics, the short vowels and phonological markers that determine pronunciation and meaning.
Restoring these marks automatically is important for text-to-speech, machine translation, and language learning \citep{habash2007arabic}, but remains challenging: text-only models \citep{alasmary2024catt} struggle with inherent ambiguity, particularly in dialectal Arabic where phonetic variation is high.
Speech signals offer a complementary source of disambiguation, as prosodic and phonetic cues can resolve cases that are irrecoverable from orthography alone.

The KSAA-2026 Shared Task \citep{ksaa2026task} presents a low-resource instance of this problem: only 2{,}327 training samples across multiple Arabic dialects, with no external data permitted.
CATT-Whisper \citep{ghannam2025catt-whisper} demonstrated that a multimodal architecture combining CATT and Whisper encoders is effective for this task, achieving strong results at NADI~2025.
We adopt this architecture and focus on making the most of the limited training data through regularization and inference-time ensembling.

\section{Background}
\label{sec:background}

\subsection{Task and Data}
\label{sec:task}

Task~2 of the KSAA-2026 Shared Task requires participants to produce fully diacritized Arabic text given speech audio and undiacritized transcripts.
Systems are evaluated using Diacritic Error Rate (DER), Word Error Rate (WER), and Sentence Error Rate (SER) \citep{fadel2019arabic}, with \textbf{WER} (with case endings, including no-diacritic positions) as the primary ranking metric.
All results in this paper are reported under this setting.

The dataset \citep{ksaa2026task} contains 2{,}327 training, 260 development, and 328 test samples spanning multiple Arabic dialects.
Each sample is a short utterance (average $\sim$7 seconds) paired with a diacritized transcript.
We filter samples where the ratio of diacritized to total characters is below 0.6, yielding 2{,}187 effective training samples.

\subsection{Related Work}
\label{sec:related}

Arabic diacritization has progressed from morphological taggers \citep{habash2007arabic} through transformer-based systems \citep{nazih2022arabic} to pretrained encoders such as AraBERT \citep{antoun2020arabert} and the character-level CATT model \citep{alasmary2024catt}.
\citet{ghannam2025catt-whisper} introduced CATT-Whisper, which fuses a CATT text encoder with Whisper \citep{radford2023robust} speech features via prefix addition, winning the NADI~2025 dialectal diacritization track.
R-Drop \citep{wu2021rdrop} penalizes divergent predictions between two dropout-masked forward passes and has been effective in low-resource settings.
MC Dropout \citep{gal2016dropout} enables ensemble-like inference from a single model by keeping dropout active at test time.
Our work applies these regularization and inference techniques to multimodal Arabic diacritization.

\section{System Overview}
\label{sec:system}

\subsection{Architecture}
\label{sec:arch}

Our system builds on CATT-Whisper.
The text encoder is a 6-layer CATT Transformer ($d\!=\!512$, 16 heads), pretrained on Arabic diacritization, predicting one of 15 diacritic classes per Arabic letter.
The speech encoder is Whisper-base \citep{radford2023robust} (6 encoder blocks, $d\!=\!512$), kept fully frozen in the primary configuration.
Fusion uses prefix addition: 1{,}500 Whisper frames are mean-pooled into 150 tokens, projected, and added to 150 dedicated prefix positions that precede the text input.
The model has $\sim$39M parameters ($\sim$19M trainable).

\subsection{Training}
\label{sec:training}

\begin{figure*}[t]
\centering
\includegraphics[width=\textwidth]{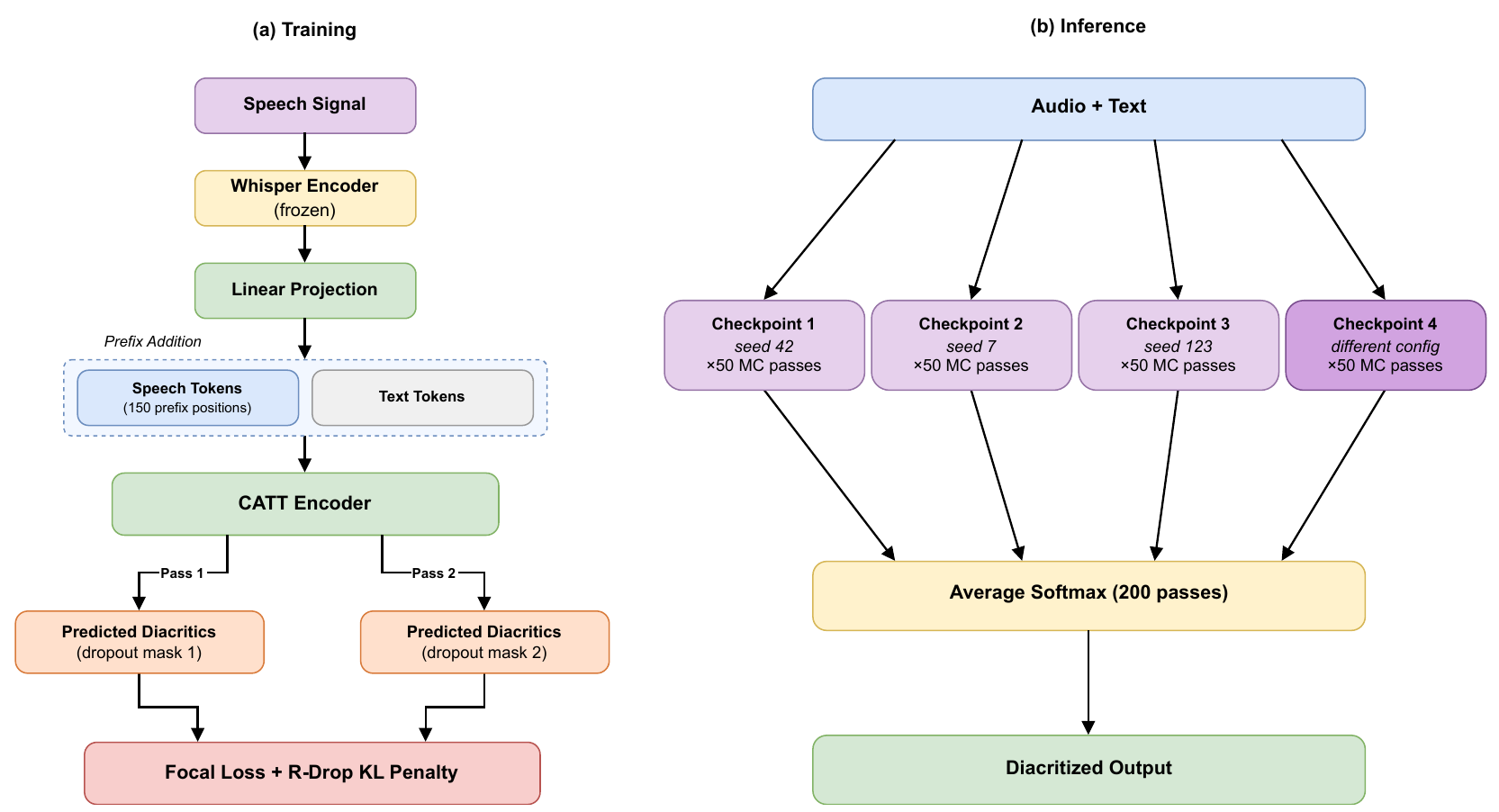}
\caption{(a)~Training: speech features from the frozen Whisper encoder are fused with text tokens via prefix addition and processed by the CATT encoder. R-Drop runs two forward passes with different dropout masks, optimizing with Focal Loss and a KL consistency penalty. (b)~Inference: four checkpoints each run 50 MC Dropout passes; the 200 softmax distributions are averaged to produce the diacritized output.}
\label{fig:system}
\end{figure*}

Figure~\ref{fig:system}(a) illustrates the training procedure.
We fine-tune with R-Drop: each input passes through the model twice with different dropout masks, and a symmetric KL divergence penalty ($\alpha\!=\!2.08$) encourages consistent predictions.
The supervised loss is Focal Loss \citep{lin2017focal} ($\gamma\!=\!0.34$) with label smoothing ($\epsilon\!=\!0.018$).
Speech embedding dropout ($p\!=\!0.09$) randomly zeros the speech representation during training, following the modality-robust training scheme of CATT-Whisper.
Audio augmentation uses SpecAugment \citep{park2019specaugment} and Gaussian noise injection.
Hyperparameters were selected using Optuna \citep{akiba2019optuna} (30 trials, 12 epochs each); Table~\ref{tab:hparams} shows the final configuration.
We use AdamW \citep{loshchilov2019adamw} with cosine learning rate decay.
We train four checkpoints: three with this configuration using seeds 42, 7, and 123, and a fourth from a separate Optuna trial with an alternative configuration (learning rate $4.7 \times 10^{-5}$, batch size 32, $\gamma\!=\!1.0$, label smoothing 0.108, and 4 unfrozen Whisper blocks after epoch 15) to increase ensemble diversity.

\begin{table}[t]
\centering
\small
\begin{tabular}{@{}lc@{}}
\toprule
\textbf{Hyperparameter} & \textbf{Value} \\
\midrule
Learning rate & $4.1 \times 10^{-6}$ \\
R-Drop $\alpha$ & 2.08 \\
Focal $\gamma$ / label smoothing & 0.34 / 0.018 \\
Weight decay & 0.098 \\
Speech emb.\ dropout & 0.09 \\
Batch size / epochs & 16 / 40 \\
Warmup / min LR factor & 3\,ep / 0.002 \\
SpecAugment (freq / time) & 10 / 63 \\
Noise injection & Gauss., SNR 10--30\,dB \\
Whisper unfrozen & 0 (fully frozen) \\
\bottomrule
\end{tabular}
\caption{Hyperparameters selected via Optuna (30 trials).}
\label{tab:hparams}
\end{table}

\subsection{Inference}
\label{sec:inference}

Figure~\ref{fig:system}(b) illustrates the inference procedure.
At inference, we keep dropout active in the CATT encoder ($p\!=\!0.1$) while LayerNorm stays in eval mode.
Each of the four models runs 50 stochastic forward passes ($4 \times 50 = 200$ total), and we average softmax probabilities across all passes before taking the argmax.

\subsection{Post-Processing}
\label{sec:postproc}

We use direct positional diacritic insertion: CATT models maintain 1:1 correspondence between Arabic letter positions and predicted diacritics.
We enforce three invariants: (1)~stripping diacritics from the output recovers the original input; (2)~diacritic count matches predictions; (3)~all letter positions are consumed.

\section{Results}
\label{sec:results}

Table~\ref{tab:leaderboard} shows the test set results compared to other participants and the shared task baselines.
Our system achieves \textbf{23.26\% WER} on the primary metric.

\begin{table}[t]
\centering
\small
\begin{tabular}{@{}lccc@{}}
\toprule
\textbf{System} & \textbf{DER\,$\downarrow$} & \textbf{WER\,$\downarrow$} & \textbf{SER\,$\downarrow$} \\
\midrule
\textbf{meshal (Ours)} & \textbf{6.87} & \textbf{23.26} & \textbf{66.16} \\
nadaadelmousa & 7.04 & 24.39 & 71.65 \\
naif\_alharthi & 7.51 & 25.34 & 73.48 \\
nahian\_abu & 8.23 & 30.37 & 80.79 \\
Hassan & 10.56 & 34.47 & 79.88 \\
omarnj10 & 27.94 & 44.05 & 98.78 \\
astral\_fate & 31.67 & 84.50 & 99.70 \\
\midrule
Baseline (FT text+ASR) & 9.91 & 31.84 & 82.93 \\
Baseline (text+ASR) & 13.50 & 40.24 & 82.32 \\
Baseline (text-only) & 17.66 & 49.85 & 91.77 \\
\bottomrule
\end{tabular}
\caption{Test set results. All metrics: with case endings, incl.\ no-diacritic positions. Ranked by WER (primary metric).}
\label{tab:leaderboard}
\end{table}

Table~\ref{tab:ablation} presents a cumulative ablation on the development set, starting from the pretrained CATT-Whisper model and progressively adding our modifications.
Fine-tuning CATT-Whisper with a standard recipe (cross-entropy loss, learning rate $10^{-5}$) gives 30.43\% WER; adding our regularized recipe (R-Drop, Focal Loss, high weight decay) reduces this to 27.18\%, and MC Dropout ensembling brings it to 26.02\%.
The majority of the gain (3.25\,pp) comes from the regularized training recipe; MC Dropout adds a further 1.16\,pp, confirming that the training recipe is the primary driver of improvement.

\begin{table}[t]
\centering
\small
\begin{tabular}{@{}lcc@{}}
\toprule
\textbf{Configuration} & \textbf{DER} & \textbf{WER} \\
\midrule
CATT-Whisper (pretrained) & 17.76 & 54.06 \\
CATT-Whisper (fine-tuned)$^\dagger$ & 8.59 & 30.43 \\
\quad + Regularized recipe$^\ddagger$ & 7.57 & 27.18 \\
\quad + 4-model MC Dropout ensemble & \textbf{7.17} & \textbf{26.02} \\
\bottomrule
\multicolumn{3}{@{}l}{\scriptsize $^\dagger$Baseline: lr=$10^{-5}$, cross-entropy loss, batch size 16, 30 epochs.} \\
\multicolumn{3}{@{}l}{\scriptsize $^\ddagger$R-Drop + Focal Loss + high weight decay (Optuna, 30 trials).}
\end{tabular}
\caption{Cumulative ablation on dev set (\%, with case endings, incl.\ no-diacritic positions).}
\label{tab:ablation}
\end{table}

\subsection{Discussion}
\label{sec:discussion}

\paragraph{Training recipe.}
We also explored architectural modifications including cross-attention fusion, CRF decoding, attention pooling, auxiliary heads, and RL fine-tuning, none of which improved over the fine-tuned CATT-Whisper baseline.
The regularized recipe yielded a 3.25\,pp WER gain (Table~\ref{tab:ablation}), suggesting that, in our experiments with 2{,}187 training samples, the optimization strategy matters more than the model architecture.
MC Dropout ensembling adds 1.16\,pp at a cost of 200 forward passes ($\sim$50$\times$ slower than a single pass); in practice, reducing the number of passes and studying the effect on accuracy could yield a more efficient trade-off.

\paragraph{Audio contribution.}
\citet{ghannam2025catt-whisper} showed that incorporating speech features significantly improves diacritization accuracy.
Consistent with their findings and the gap between the text-only and audio-equipped baselines in Table~\ref{tab:leaderboard}, we observed a similar pattern when fine-tuning CATT-Whisper without speech features, further confirming the importance of audio input for this task.

\paragraph{Qualitative example.}
Table~\ref{tab:examples} shows a dev set example where our system correctly diacritizes a complex sentence with case endings and shaddas.

\begin{table}[t]
\centering
\small
\begin{tabular}{@{}lp{0.72\columnwidth}@{}}
\toprule
Input & \ar{الظاهر أنه لا خلاف في الحقيقة للاتفاق على امتناع إدراك حقيقة الذات} \\
Ours & \ar{الظَّاهِرُ أَنَّهُ لَا خِلَافَ فِي الْحَقِيقَةِ لِلِاتِّفَاقِ عَلَى امْتِنَاعِ إِدْرَاكِ حَقِيقَةِ الذَّاتِ} \\
Gold & \ar{الظَّاهِرُ أَنَّهُ لَا خِلَافَ فِي الْحَقِيقَةِ لِلِاتِّفَاقِ عَلَى امْتِنَاعِ إِدْرَاكِ حَقِيقَةِ الذَّاتِ} \\
\bottomrule
\end{tabular}
\caption{Dev set example.}
\label{tab:examples}
\end{table}

\section{Conclusion}
\label{sec:conclusion}

We presented our system for KSAA-2026 Task~2, which fine-tunes CATT-Whisper with R-Drop, Focal Loss, and Optuna-optimized hyperparameters, and uses MC Dropout ensembling at inference.
At this data scale, training regularization yielded larger gains than any of the architectural modifications we explored.
Future work could disentangle the individual contributions of each regularization component and examine per-dialect performance to identify remaining challenges.

\section*{Acknowledgements}

We thank Thaka for supporting this work, KSAA for organizing the shared task, and the Abjad~AI team for open-sourcing the CATT-Whisper model.

\section*{References}
\label{sec:reference}

\bibliographystyle{lrec2026-natbib}
\bibliography{references}

\end{document}